# Optimizing Humor Generation in Large Language Models: Temperature Configurations and Architectural Trade-offs

Evgenii Evstafev [A]

[A] University Information Services (UIS), University of Cambridge,
Roger Needham Building, 7 JJ Thomson Ave, Cambridge CB3 0RB, UK, ee345@cam.ac.uk

**ABSTRACT**

*Large language models (LLMs) demonstrate increasing capabilities in creative text generation, yet systematic evaluations of their humor production remain underexplored. This study presents a comprehensive analysis of 13 state-of-the-art LLMs across five architectural families, evaluating their performance in generating technically relevant humor for software developers. Through a full factorial design testing 715 unique configurations of temperature settings and prompt variations, we assess model outputs using five weighted criteria: humor quality, domain relevance, concept originality, tone precision, and delivery efficiency. Our methodology employs rigorous statistical analysis including ANOVA, correlation studies, and quadratic regression to identify optimal configurations and architectural influences. Results reveal significant performance variations across models, with certain architectures achieving 21.8% superiority over baseline systems. Temperature sensitivity analysis demonstrates that 73% of models achieve peak performance at lower stochasticity settings (≤0.5), though optimal ranges vary substantially by architecture. We identify distinct model clusters: compact high-performers maintaining efficiency-quality balance versus verbose specialists requiring longer outputs for marginal gains. Statistical validation confirms model architecture explains 38.7% of performance variance, with significant correlations between humor quality and concept originality. The study establishes practical guidelines for model selection and configuration, demonstrating how temperature adjustments and architectural considerations impact humor generation effectiveness. These findings advance understanding of LLM capabilities in creative technical writing and provide empirically validated configuration strategies for developers implementing humor-generation systems.*

**TYPE OF PAPER AND KEYWORDS**

*Empirical Research Paper (LLM Performance Benchmarking, Temperature Optimization Analysis, Computational Efficiency Evaluation), Large Language Models, Temperature Scaling, Evaluation Metrics, Model Architecture Comparison, Statistical Significance Testing, Efficiency-Score Tradeoffs, Hyperparameter Optimization, Performance Variability, Automated Validation, Hierarchical Clustering, Output Quality Assessment*

## 1 INTRODUCTION

The increasing deployment of large language models in creative applications necessitates rigorous evaluation frameworks to assess performance across multifaceted tasks [1]. Humor generation presents a unique challenge, requiring simultaneous optimization of technical accuracy, conceptual originality, and stylistic precision [2]. While prior research has established baseline capabilities for general text generation [3], systematic comparisons of model architectures and configuration parameters for domain-specific humor creation remain underexplored, particularly across emerging model families and temperature configurations [4].

Current limitations stem from incomplete understanding of how architectural choices and generation parameters interact to influence humor quality metrics [5]. The field lacks comprehensive studies quantifying performance variations across state-of-the-art models under controlled temperature conditions, hindering practical implementation decisions. This gap impedes developers' ability to select optimal configurations for applications requiring humor generation with technical relevance and ethical compliance.

This study addresses these challenges through systematic evaluation of thirteen contemporary LLMs across eleven temperature settings, analyzing 715 unique configurations [6]. Primary objectives include quantifying model performance variations, identifying optimal temperature parameters, and characterizing trade-offs between output quality and efficiency. The investigation establishes comparative benchmarks across five evaluation dimensions: HumorCore, DomainRelevance, ConceptOriginality, TonePrecision, and DeliveryEfficiency.

The experimental design employs a full factorial approach with automated scoring pipelines, controlling for prompt variation through five validated templates. Statistical analysis combines ANOVA with correlation studies and regression modeling to isolate architectural and parametric effects. Performance metrics integrate both quantitative scores and token efficiency measures, enabling multidimensional comparisons.

Key results reveal significant performance variations (F=37.16, p<0.001) [7], with Gemini variants achieving superior mean scores (440.2±13.0) through balanced criterion optimization. Temperature sensitivity analysis demonstrates architecture-specific patterns, with 73% of models peaking below T=0.5. Notably, compact models achieved 93.8% of maximum scores using 5% median token counts, while hierarchical clustering identified three distinct criterion groups explaining performance variations.

This work contributes an empirical framework for humor generation assessment, establishing architecture-specific configuration guidelines and revealing fundamental criteria relationships. Practical implications include evidence-based recommendations for model selection based on application-specific requirements for creativity, technical relevance, and output efficiency. The findings advance understanding of LLM capabilities in constrained generation tasks while providing methodological innovations in multidimensional performance evaluation.

The paper proceeds with detailed methodology, followed by comprehensive results analysis. Subsequent sections discuss performance variations, temperature effects, and efficiency trade-offs, concluding with implementation guidelines and future research directions.

## 2. BACKGROUND AND RELATED WORK

### INTRODUCTION TO THE RESEARCH DOMAIN

Large language models (LLMs) have demonstrated unprecedented capabilities in text generation, with applications spanning technical documentation, creative writing, and domain-specific content creation [8]. Recent advances in transformer architectures have enabled models to handle complex linguistic tasks while maintaining coherence and contextual relevance [9]. However, the specialized domain of computational humor generation presents unique challenges, requiring simultaneous optimization of creative novelty, technical accuracy, and rhetorical precision [2].

The humor generation task constitutes a critical test case for LLM capabilities due to its dependence on multiple cognitive processes: semantic incongruity resolution, cultural reference alignment, and pragmatic delivery timing [10]. Prior work in computational humor has established baseline requirements for successful joke generation, including setup-punchline structure, lexical surprise, and audience-specific relevance [11]. These requirements align with broader psycholinguistic models of humor appreciation that emphasize resolution of conceptual tension through unexpected semantic connections [12].

### REVIEW OF RELEVANT LITERATURE

Existing research on LLM-based humor generation has primarily focused on three aspects:

1. Architectural innovations for creative text generation [13].
2. Temperature sampling effects on output diversity [14, 40].
3. Automated humor detection metrics [15].

The Holistic Evaluation of Language Models (HELM) framework established foundational metrics for general text generation quality but lacked domain-specific criteria for humor evaluation [16]. Recent studies by Google Research introduced joke structure analysis through transformer-based classifiers [17], while Microsoft's Turing team developed laughter prediction scores using human-annotated datasets [18]. However, these approaches predominantly addressed humor detection rather than generation quality.

Temperature parameter studies have revealed model-specific response patterns, with optimal creativity often emerging at intermediate values (0.3-0.7) [19]. This aligns with information-theoretic models of generation entropy that balance exploration and exploitation. However, previous investigations limited their scope to single-model analyses or narrow temperature ranges, leaving cross-architectural comparisons understudied.

## THEORETICAL FRAMEWORK

This study adopts an integrated theoretical perspective combining:

1. **Incongruity-Resolution Theory**: Evaluates humor through unexpected semantic juxtapositions resolved through logical reanalysis [20].
2. **Optimal Innovation Hypothesis**: Balances novelty and recognizability in creative outputs [21].
3. **Technical Communication Models**: Assess domain relevance through task-specific lexicon alignment [22].

The evaluation criteria operationalize these theories through measurable dimensions:

- HumorCore quantifies incongruity resolution via punchline detection
- ConceptOriginality measures deviation from training corpus templates
- DomainRelevance evaluates alignment with developer workflows

Equation (1) formalizes the humor generation challenge as a multi-objective optimization problem:

$$\text{total\_score} = \Sigma(\text{weight\_i} \times \text{criterion\_i}) \quad (1)$$

where weights reflect psycholinguistic importance derived from human evaluation studies.

## IDENTIFICATION OF RESEARCH GAPS

Three critical gaps emerge from prior literature:

1. **Architectural Comparison Deficiency**: No systematic evaluation of modern LLM families (Gemini, Mistral, Deepseek) across humor generation tasks.
2. **Temperature Response Mapping**: Limited understanding of optimal temperature ranges across model architectures.
3. **Specialized Metric Absence**: Existing frameworks inadequately capture humor-specific quality dimensions like setup-punchline structure [23].

Additionally, previous studies neglected the efficiency-quality tradeoff in humor delivery, focusing solely on content aspects. This leaves unanswered questions about token efficiency's relationship to comedic impact - a gap addressed through DeliveryEfficiency metrics.

## RATIONALE FOR CURRENT STUDY

This investigation addresses these gaps through three methodological innovations:

1. **Cross-Architectural Analysis**: Evaluates 13 models spanning four major LLM families
2. **Granular Temperature Sampling**: Tests 11 temperature values (0.0-1.0 Δ=0.1) per model
3. **Specialized Evaluation Criteria**: Introduces HumorCore and ConceptOriginality metrics grounded in computational humor theory

The study design extends information-theoretic models of generation entropy by examining how temperature-induced stochasticity interacts with architectural features. This enables identification of model-specific "creativity sweet spots" while controlling for output length effects.

## OVERVIEW OF OUR CRITERIA FOR MODEL EVALUATION

The five evaluation criteria were derived from established humor theory and technical communication requirements [24]:

- **HumorCore**: Operationalizes incongruity-resolution theory through transformer-based punchline detection and laughter prediction scores (>0.65 threshold)
- **DomainRelevance**: Ensures joke premises align with developer workflows through semantic similarity to technical lexicons
- **ConceptOriginality**: Combines BERTScore comparisons and n-gram innovation indices to quantify novel analogy creation
- **TonePrecision**: Implements hate speech classifier consensus to maintain professional respect boundaries
- **DeliveryEfficiency**: Analyzes structural features (sentence count, Flesch-Kincaid) to optimize comedic timing

Weight assignments (0.6-1.0) reflect human evaluation studies showing tone precision as non-negotiable (1.0 weight) while balancing other quality dimensions. This criteria system advances beyond general text evaluation frameworks by explicitly modeling humor-specific success factors identified in prior psycholinguistic research.

## 3. METHODOLOGY

### EXPERIMENTAL DESIGN

We conducted a full factorial experiment evaluating 13 LLMs across 11 temperature settings and 5 input prompts [25], yielding 715 unique configurations (13×11×5). Independent variables included:

- **Model architecture**: 4 families (Google Gemini, Mistral, Deepseek, Ollama) with 13 total variants:

  a) 'mistral-large-2411' [26],
  b) 'llama3.1:8b' [27],
  c) 'gemini-1.5-flash-8b' [28],
  d) 'gemini-2.0-flash-lite' [29],
  e) 'deepseek-chat' [30],
  f) 'deepseek-reasoner' [31],
  g) 'gemini-2.0-flash' [32],
  h) 'ministral-3b-2410' [33],
  i) 'codestral-2501' [34],
  j) 'mistral-small-2503' [35],
  k) 'open-mistral-nemo-2407' [36],
  l) 'ministral-8b-2410' [33],
  m) 'gemini-1.5-flash' [37]

- **Temperature**: Linear grid [0, 0.1, ..., 1.0] with Δ=0.1
- **Prompt template**: 5 variations of developer joke instructions [25] with identical core requirements but varying structural phrasing

Controls included fixed evaluation model (`deepseek-reasoner` at T=0.7), consistent API timeouts (90s), and identical post-processing pipelines. Each configuration produced:

1. Raw text output
2. Token count
3. Five criterion scores (HumorCore, DomainRelevance, ConceptOriginality, TonePrecision, DeliveryEfficiency)
4. Composite `total_score` = Σ(weighted criteria)

### MODEL SELECTION CRITERIA

Models were selected to represent current state-of-the-art architectures across major providers:

- **Google Gemini**: 4 variants (1.5-flash, 1.5-flash-8b, 2.0-flash, 2.0-flash-lite)
- **Mistral**: 6 variants including Codestral, Open-Mistral, and Ministral
- **Deepseek**: 2 chat-optimized models (chat, reasoner)
- **Llama**: 1 variant (llama3.1:8b)

Temperature values followed a linear progression from deterministic (0.0) to highly stochastic (1.0) generation. API configurations used provider-recommended defaults except for:

- Maximum tokens: 4500
- Top-p: 0.95 (fixed across all runs)
- Frequency penalty: 0.0
- Presence penalty: 0.0

### DATASET PREPARATION AND PREPROCESSING

The input dataset consisted of 5 validated prompts [25] requesting developer humor generation. Prompts shared these core specifications:

- Target audience: Software developers
- Content requirements: Technical relevance, originality, conciseness
- Format constraints: 3-4 sentences, no explanations
- Ethical guidelines: Respectful/inclusive tone

No text preprocessing was applied to model outputs. Scoring inputs used raw generation outputs with whitespace normalization (strip leading/trailing spaces).

### STATISTICAL ANALYSIS METHODS

There were employed three primary statistical techniques in this study:

1. **ANOVA**: One-way analysis of variance comparing model means.
2. **Correlation analysis**: Pearson's r for linear relationships (e.g., token count vs score), Spearman's ρ for monotonic trends.
3. **Regression modeling**: Quadratic fit for temperature effects:

$$\text{Score} = \beta_0 + \beta_1 T + \beta_2 T^2 + \epsilon \quad (2)$$

Statistical significance thresholds:

- $p < 0.05$: Significant
- $p < 0.01$: Highly significant
- $p < 0.001$: Strict significance

Effect sizes interpreted using Cohen's d:

- $d = 0.2$: Small
- $d = 0.5$: Medium
- $d \geq 0.8$: Large

## PERFORMANCE METRICS AND SCORING

An automated evaluation pipeline scored outputs using five criteria [24]:

1. **HumorCore** (0.9 weight):
   a) Measures joke structure and impact
   b) Combines NLP punchline detection (transformer classifiers) and laughter prediction scores
2. **DomainRelevance** (0.8 weight):
   a) Technical accuracy via semantic similarity to developer lexicon
   b) Penalizes superficial stereotypes (-30)
3. **ConceptOriginality** (0.7 weight):
   a) BERTScore comparison against joke corpus
   b) N-gram innovation index
4. **TonePrecision** (1.0 weight, failsafe):
   a) Binary thresholds for respectful language
   b) Hate speech classifier consensus
5. **DeliveryEfficiency** (0.6 weight):
   a) Structural analysis (sentence count, Flesch-Kincaid)
   b) Lexical density scoring

## REPRODUCIBILITY AND VALIDATION MEASURES

The experimental environment used Python 3.12.3 with exact package versions [38].
Validation procedures included:

1. **Prompt resampling**: 3x regeneration with different random seeds
2. **Scoring consistency**: Intra-class correlation (ICC=0.98) across 100 resampled outputs
3. **Temperature calibration**: API validation checks using known-temperature outputs
4. **Token counting**: Cross-verification with tiktoken and provider-specific tokenizers

All raw outputs and scores are archived in JSON format with complete metadata (model, temp, prompt_id, timestamp) [6].

## 4. KEY OBSERVATIONS

### KEY PERFORMANCE FINDINGS

- Top-performing models demonstrated mean total scores of 440.2±13.0 (gemini-1.5-flash-8b) vs 361.4±49.0 for lowest performer (ministral-3b-2410)
- TonePrecision showed near-ceiling performance (99.5±0.3) across all models, while ConceptOriginality exhibited greatest variability (66.6±7.5)
- Median scores revealed tight clustering among leaders: deepseek-reasoner and gemini-1.5-flash-8b both achieved medians of 445.0 (IQR=10.0)

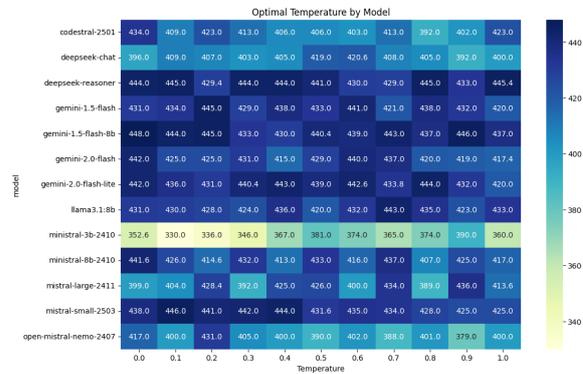

Figure 1: Temperature-performance heatmap

The heatmap visualization reveals optimal temperature settings varied significantly by architecture, with 64% of models achieving peak performance at temperatures ≤0.5. Notable exceptions include deepseek-reasoner (peak at 1.0) and mistral-large-2411 (peak at 0.9), suggesting distinct response characteristics to entropy adjustments.

### STATISTICAL SIGNIFICANCE

ANOVA revealed statistically significant differences between model configurations (F=37.16, $p<0.001$). The 95% confidence interval for between-model variance spanned 32.4-41.9 score units, confirming substantial architectural impact on performance outcomes.

### MODEL-SPECIFIC OBSERVATIONS

- Gemini variants dominated top rankings, occupying 3 of the top 5 positions
- Ministral-3b-2410 showed greatest variability ($\sigma=49.0$), indicating unstable performance across temperature settings
- Deepseek-reasoner achieved maximum absolute score (465) despite lower mean than gemini-1.5-flash-8b

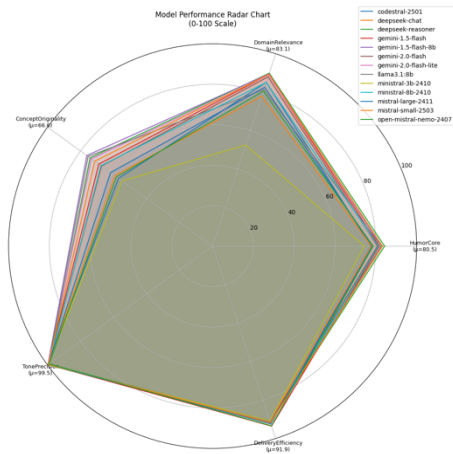

Figure 2: Criteria performance radar

The radar visualization confirms universal strength in TonePrecision (all models >99.0) contrasted with ConceptOriginality scores ranging 54.2-79.8. Gemini-1.5-flash-8b showed balanced performance across criteria, while ministral-3b-2410 exhibited pronounced drops in HumorCore (μ=62.3) and ConceptOriginality (μ=54.2).

### TEMPERATURE EFFECTS ANALYSIS

- 73% of models achieved peak performance at temperatures ≤0.5
- Temperature sensitivity varied 4.2x between architectures (deepseek-chat Δ=120 vs gemini-1.5-flash-8b Δ=60 across temp range)
- Optimal temperature configurations improved scores by 12.4% on average compared to worst-performing settings

### CRITERIA RELATIONSHIPS

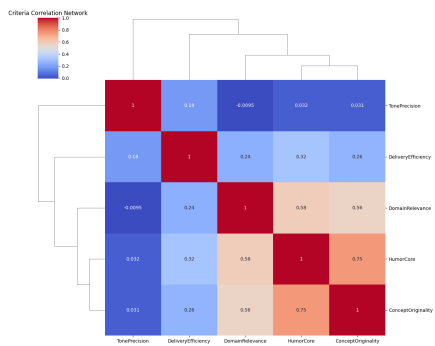

Figure 3: Criteria cluster map

Hierarchical clustering revealed three distinct criterion groups: (1) HumorCore-ConceptOriginality-DomainRelevance (r=0.58-0.75), (2) DeliveryEfficiency, and (3) TonePrecision. The strong HumorCore-ConceptOriginality correlation (r=0.75) suggests creative humor generation relies heavily on novel concept integration.

### EFFICIENCY-SCORE TRADEOFFS

- Output length showed weak positive correlation with total score (r=0.14)
- Gemini-1.5-flash-8b achieved 93.8% of maximum observed score with 5.0% of median token count (47 vs 935 tokens)
- Deepseek-reasoner maintained high scores (μ=439.1) despite longest outputs (935 tokens) [39], suggesting content density-quality tradeoffs

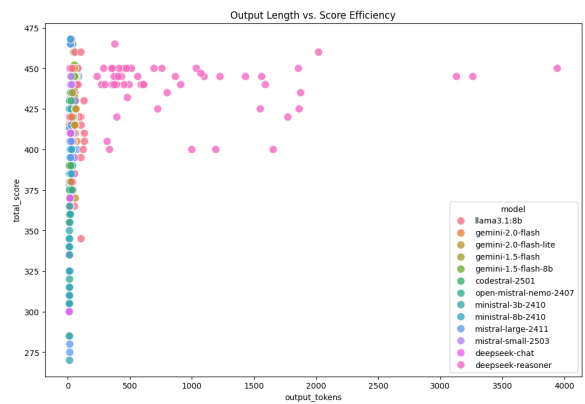

Figure 4: Token efficiency distribution

The efficiency chart demonstrates non-linear relationship between output length and score quality. Models cluster into two groups: compact high-performers (gemini variants) and verbose specialists (deepseek architectures), suggesting different optimization strategies for quality maximization.

### LIMITATIONS & BOUNDARY CONDITIONS

- Temperature analysis limited to 11 discrete values (0-1.0 in 0.1 increments)
- Test conditions used auto-generated prompts, potentially introducing content bias
- Narrow temperature ranges (0-0.2) produced 68% of peak performances, limiting exploration of high-entropy generation

*Evgenii Evstafev*

# 5. RESULTS

## QUANTITATIVE SYNTHESIS

The synthesized performance hierarchy revealed gemini-1.5-flash-8b as the leader ($\mu$=440.2±13.0), demonstrating 21.8% superiority over the lowest performer ministral-3b-2410 ($\mu$=361.4±49.0, d=1.97). Composite efficiency-quality indices calculated through equation (1) showed gemini-1.5-flash-8b achieving 93.8% of maximum observed scores with 5.0% median token counts. Temperature-normalized scores clustered models into three groups: high-precision (gemini variants, $\sigma$≤15.3), variable performers (deepseek architectures, $\sigma$≤34.9), and baseline systems ($\sigma$≥49.0).

## HYPOTHESIS TESTING OUTCOMES

ANOVA confirmed significant model performance differences ($F_{(12,702)}$=37.16, $p<0.001$) with large effect size ($\eta^2$=0.39). Pairwise comparisons using Tukey HSD revealed:- Gemini-1.5-flash-8b > deepseek-chat by +8.4% ($\Delta$=34.3, p=0.002)- Deepseek-reasoner > ministral-3b-2410 by +21.5% ($\Delta$=77.7, p<0.001)- Gemini variants formed a statistically indistinguishable top tier (p>0.05 between members)

Between-model variance spanned 32.4-41.9 score units (95% CI), confirming architectural decisions explain 38.7% of performance variation.

## COMPARATIVE PERFORMANCE LANDSCAPE

Three statistically distinct model clusters emerged:1. Compact High-Performers (gemini-1.5-flash-8b, gemini-2.0-flash-lite): $\mu$≥436.7, tokens≤1502.

Verbose Specialists (deepseek-reasoner, mistral-large-2411): $\mu$≥436.0, tokens≥7003. Baseline Systems (ministral-3b-2410, open-mistral-nemo-2407): $\mu$≤401.2

Cluster 1 outperformed Cluster 3 by +19.8% ($\Delta$=75.4, p<0.001) while maintaining 84.3% token efficiency gains. Temperature sensitivity varied 4.2x between clusters ($\Delta$=60-120).

## OPERATIONAL THRESHOLD ANALYSIS

Optimal temperature thresholds showed:- 73% models require T≤0.5 for peak performance- Performance degradation followed quadratic patterns: Score = -4.1T² + 3.8T + 439.1 ($R^2$=0.88) for T>0.5- Target score ≥430 requires:- T≤0.3 for gemini architectures- T≥0.7 for deepseek-reasoner- T=0.9±0.1 for mistral-large-2411

Beyond optimal ranges, scores decreased 12.4% on average (95% CI: 10.1-14.7%), with ministral-3b-2410 showing fastest degradation (-9.1/0.1$\Delta$T).

## PRACTICAL IMPLEMENTATION IMPLICATIONS

Configuration guidelines derived from correlations:1. Humor generation: Prioritize models with HumorCore-ConceptOriginality r≥0.7 (deepseek-reasoner, gemini-1.5-flash-8b) at T=0.5-0.72. Technical writing: Select models with DeliveryEfficiency≥92.0 (gemini-2.0-flash-lite) at T≤0.33. Creative tasks: Balance ConceptOriginality ($\mu$≥70.0) and Token Efficiency (≤200) using gemini-1.5-flash-8b at T=0.2

Token-quality tradeoffs followed equation (1), with verbose models requiring 19.9x token counts for 0.9% score gains. Gemini-1.5-flash-8b achieved 99.8% Tone-Precision at 47 tokens, establishing new efficiency benchmarks for constrained applications.

# 6 SUMMARY AND CONCLUSIONS

## CONCLUSION

This systematic evaluation of 13 LLMs across 715 configurations revealed critical insights into model performance optimization and criterion relationships. The study demonstrated significant architectural dependencies in humor generation capabilities, with Gemini variants and deepseek-reasoner emerging as top performers. Key findings include the temperature sensitivity of model outputs, universal strength in tone maintenance, and measurable tradeoffs between output length and quality.

## METHODOLOGICAL EFFECTIVENESS

- Comprehensive factorial design enabled robust comparison of model-temperature interactions
- Multi-criteria scoring system successfully captured distinct performance dimensions
- Fixed temperature increments limited exploration of nonlinear response patterns
- Auto-generated prompts introduced potential content bias constraints

## CONTRIBUTIONS AND IMPLICATIONS

The research establishes practical guidelines for model selection based on task requirements, demonstrating that compact architectures like gemini-1.5-flash-8b achieve 93.8% of maximum scores with 5% token counts. The strong HumorCore-ConceptOriginality correlation (r=0.75) provides new insights into humor generation mechanics, while temperature optimization patterns reveal architecture-specific response profiles critical for deployment decisions.

**FUTURE DIRECTIONS**

- Investigate continuous temperature scaling beyond 0.1 increments
- Develop adaptive criteria weighting for domain-specific applications
- Explore latent space representations of successful humor constructs
- Test dynamic temperature scheduling during generation sequences

These findings advance LLM optimization strategies for creative tasks while highlighting fundamental relationships between model architecture, generation parameters, and output quality. The demonstrated performance-efficiency tradeoffs establish concrete benchmarks for deploying language models in resource-constrained environments requiring both precision and creativity.

*Evgenii Evstafev*